\documentclass[letterpaper]{article} 
\usepackage{aaai2026}  
\usepackage{times}  
\usepackage{helvet}  
\usepackage{courier}  
\usepackage[hyphens]{url}  
\usepackage{graphicx} 
\usepackage{natbib}  
\usepackage{caption} 
\usepackage{algorithm}
\usepackage{algorithmic}
\usepackage{amsmath, amsfonts}
\usepackage{subcaption}
\usepackage{booktabs}
\usepackage{newfloat}
\usepackage{listings}
\DeclareCaptionStyle{ruled}{labelfont=normalfont,labelsep=colon,strut=off} 
\floatstyle{ruled}
\newfloat{listing}{tb}{lst}{}
\floatname{listing}{Listing}
\title{StyleProtect: Safeguarding Artistic Identity in Fine-tuned Diffusion Models}
\author{
    Qiuyu Tang,
    Joshua Krinsky,
    Aparna Bharati\\
}
\affiliations{
    Department of Computer Science and Engineering\\

    Lehigh University\\
    Bethlehem, PA 18015 USA\\
    \{qit220, jpk322, apb220\}@lehigh.edu
}

\begin{document}

\maketitle

\begin{abstract}
\label{sec:abstract}
The rapid advancement of generative models, particularly diffusion-based approaches, has inadvertently facilitated their potential for misuse. Such models enable malicious exploiters to replicate artistic styles that capture an artist’s creative labor, personal vision, and years of dedication in an inexpensive manner. This has led to a rise in the need and exploration of methods for protecting artworks against style mimicry. Although generic diffusion models can easily mimic an artistic style, finetuning amplifies this capability, enabling the model to internalize and reproduce the style with higher fidelity and control.
We hypothesize that certain cross-attention layers exhibit heightened sensitivity to artistic styles. Sensitivity is measured through activation strengths of attention layers in response to style and content representations, and assessing their correlations with features extracted from external models. Based on our findings, we introduce an efficient and lightweight protection strategy, \textit{StyleProtect}, that achieves effective style defense against fine-tuned diffusion models by updating only selected cross-attention layers.
Our experiments utilize a carefully curated artwork dataset based on WikiArt, comprising representative works from 30 artists known for their distinctive and influential styles and cartoon animations from the Anita dataset. The proposed method demonstrates promising performance in safeguarding unique styles of artworks and anime from malicious diffusion customization, while maintaining competitive imperceptibility.
\end{abstract}

\section{Introduction}
\label{sec:introduction}



The widespread adoption of generative models~\cite{li2024autoregressive,goodfellow2014generative,radford2015unsupervised,Richardson_2021_CVPR}, especially diffusion-based models~\cite{ho2020denoising, song2021denoising, rombach2022high, kumari2023multi}, has significantly transformed the landscape of image synthesis, lowering technical barriers and enabling users without formal artistic training to produce photorealistic visual content. While greater accessibility to diffusion-based models has broadened participation in visual content creation, it has raised concerns about the unauthorized and unacknowledged use of established artistic styles. For instance, an individual fine-tuned a diffusion model on a small set of artworks created by the American illustrator Hollie Mengert using DreamBooth~\cite{ruiz2023dreambooth}, a publicly accessible fine-tuning technique. The resulting model was then used to generate new images that closely mimicked her visual style which were publicly shared without attribution~\cite{jackson2022invasive}.

\begin{figure}[t]
    \centering
    \includegraphics[width=\linewidth]{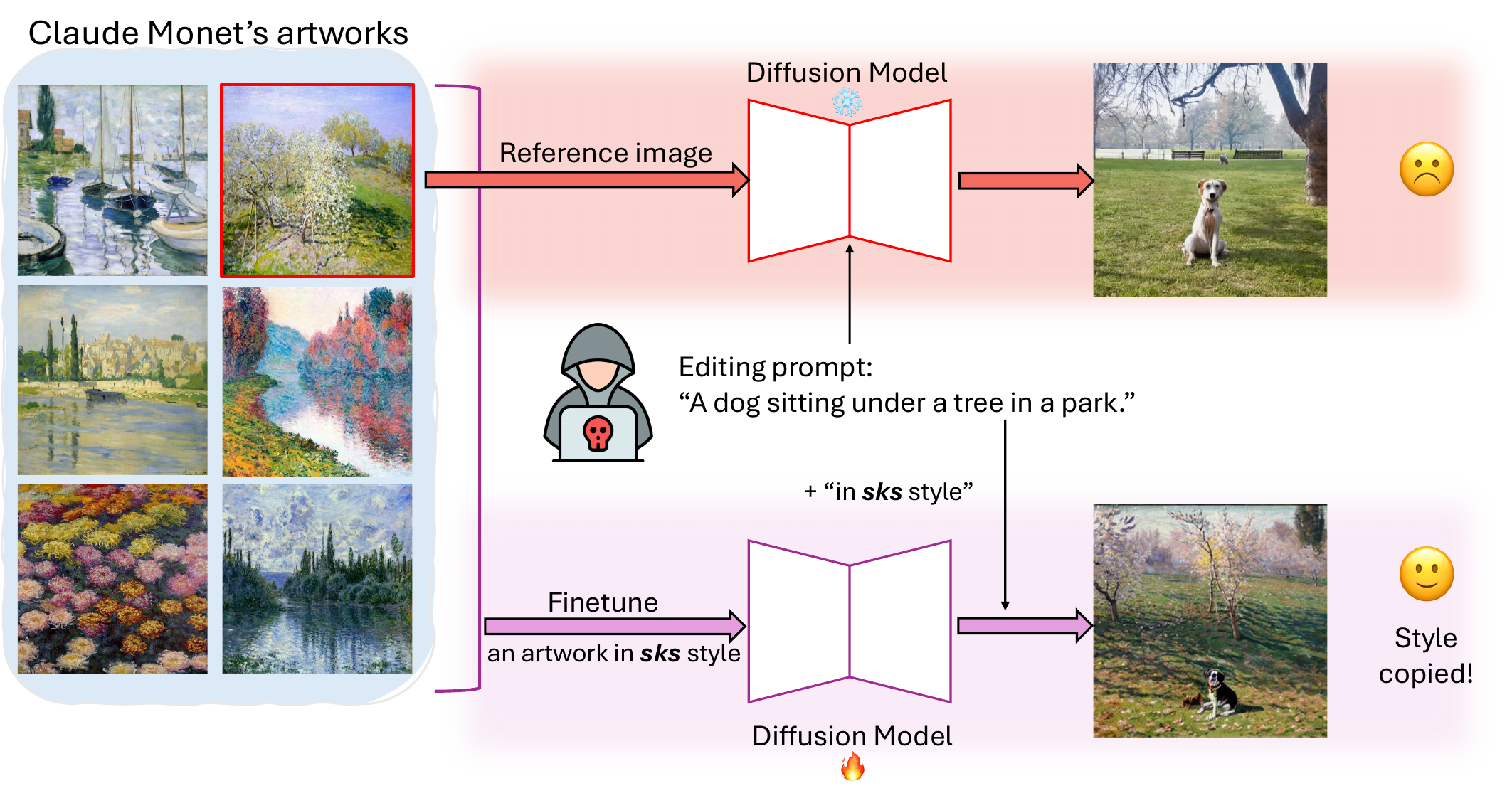}
    \caption{Fine-tuning enables more faithful,  consistent style replication than inference-based methods. 
    Top: A frozen model uses a Monet painting as reference to generate “a dog sitting under a tree in a park.”
    Bottom: A model is fine-tuned on Monet’s works with the style bound to the token “sks,” enabling “a dog sitting under a tree in a park in \textit{sks} style” to produce consistent stylized outputs.}
    \label{fig:inferenceVSfinetune}
\end{figure}

Artworks hold cultural, historical, and emotional value, offering insights into past lives and practices while connecting artists with audiences. This significance underscores the need to protect them from AI mimicry.
Style mimicry in this context refers to the ability of diffusion models to replicate an artist’s unique visual style by learning from reference artworks. This capability raises significant ethical and economic concerns.
Malicious individuals may exploit AI models to mass-produce counterfeit artworks for profit or recognition.
The victory of AI-generated images at the Colorado State Fair's annual art competition and other prestigious venues ignited a debate about the nature of art~\cite{nyt_ai_art_2022, nyt_ai_generated_art_2022}.
This victory challenged traditional notions of art, raising concerns that AI-generated work could undermine original artists' rights, devalue human creativity, and damage trust in the creative community.

To safeguard original images from the risks associated with generative models, researchers have proposed protection measures~\cite{salman2023raising, shan2023glaze, liang2023mist, cui2025diffusionshield, cui2025ft}. Prior study~\cite{tang2025watermarks} has explored both watermark-based and perturbation-based methods. Watermarks are designed to help trace the origin of generated images, whereas perturbations aim to disrupt the image generation process. Degrading the performance of generative models offers an effective measure of making artistic styles difficult to learn. Moreover, slight perturbations typically do not affect human perception of the images. Hence, we focus on perturbation-based protection methods.

Recognizing the risks of diffusion models,~\cite{salman2023raising} introduced PhotoGuard, a perturbation-based method to deter malicious editing. Building on this, some recent works~\cite{liang2023mist, shan2023glaze, ahn2024nearly} have enhanced protection for artwork and others are specialized to defend against biometric privacy violations~\cite{van2023anti, xu2024perturbing, liu2024metacloak}. However, fine-tuning techniques~\cite{gal2022image, ruiz2023dreambooth, hu2022lora, kumari2023multi} now pose new threats by enabling exploiters to generate high-quality stylistic counterfeits (see Figure~\ref{fig:inferenceVSfinetune})
and protection against such mimicry remains underexplored. 

Our work aims to defend artistic styles from unauthorized personalization and mimicry enabled by diffusion-based fine-tuning. Artistic style embodies an artist’s distinctive creative identity—an expression refined over years and inherently difficult for humans to replicate. Unlike traditional image protection techniques such as watermarking or content obfuscation, which safeguard ownership or content, style protection targets the intangible yet deeply personal elements of visual art—brush strokes, color palettes, compositional structure, and rhythmic patterns.
To this end, we introduce \textit{StyleProtect}, an adversarial perturbation-based defense that disrupts unauthorized style replication. Our perturbations are strategically and efficiently optimized by targeting style-sensitive cross-attention layers within the diffusion model.
In summary, our contributions are:

\begin{enumerate}
    \item Style-based Cross-attention Layer Selection: We identify and validate that certain cross-attention layers in diffusion models are more sensitive to artistic style. 
    \item Perturbation-Based Defense: We propose a novel and efficient perturbation-based method, \textit{StyleProtect}, that restricts updates to style-sensitive cross-attention layers. 
    \item Curated Artistic Dataset and Comprehensive Evaluation: We curate a high-quality artwork dataset featuring diverse and distinctive styles. Extensive experiments demonstrate the effectiveness, imperceptibility, robustness, and efficiency of our method, outperforming existing protection baselines. We also assess the effectiveness on cartoon animations from the Anita Dataset.
\end{enumerate}

This paper is organized as follows. 
We begin by reviewing background literature on diffusion-based generation, current protection methods, and the role of diverse artistic descriptors. Next, we introduce our proposed method, \textit{StyleProtect}, outlining the selection of style-sensitive layers and the design of targeted perturbations. Finally, we detail our experimental setup, present results and ablation studies, and 
conclude with a discussion of key findings, limitations and future research.

\section{Related Work}
\label{sec:relatedwork}


In this section, we begin by revisiting the pressing issue of style mimicry and explore how various models interpret and represent artistic style. We then describe the generative process of diffusion models, with a focus on their applications in style editing. Finally, we review recent works that protect visual assets from such manipulations. 

\subsection{Style Representation and Description}

Understanding and disentangling artistic style from artwork has long been a fundamental challenge in computer vision. Early algorithms~\cite{gibson1966senses, lun2015elements, sablatnig1998hierarchical, silva2021automatic, yao2009characterizing} attempted to explicitly encode style by decomposing images into handcrafted, low-level features such as shape descriptors, color histograms, and texture patterns. These methods treated style as a set of predefined visual attributes, often resulting in limited expressiveness and flexibility.

Instead of manually designing features, modern approaches~\cite{gatys2016image, wynen2018unsupervised, jing2019neural, li2017demystifying, Huang_2018_ECCV, chu2018image, xu2014architectural, matsuo2016cnn, ruta2021aladin} focus on using deep learning models to implicitly represent style. \cite{gatys2016image} introduced the use of Gram matrices computed from feature activations in pretrained convolutional neural networks (CNNs) to capture the texture and style information of an image. This method laid the groundwork for neural style transfer~\cite{jing2019neural, li2017demystifying, zhang2013style, Huang_2018_ECCV}. Beyond transfer applications, it also sparked the development of style classification~\cite{chu2018image, xu2014architectural} and artwork retrieval systems~\cite{matsuo2016cnn, ruta2021aladin}. 

More recent studies~\cite{somepalli2024measuring, zhang2023inversion, liu2024unziplora, ahn2024dreamstyler, patashnik2021styleclip, hu2024diffusest} have further explored disentangled representations of style and content with the help of multimodal models like CLIP~\cite{radford2021learning}, enabling flexible manipulation and transfer of artistic attributes. The integration of Vision Transformer (ViT)~\cite{vaswani2017attention} within CLIP 
also fosters a more robust understanding of global image features and concepts through their patch-to-patch attention mechanism.
CSD~\cite{somepalli2024measuring} encodes style similarity by employing contrastive learning to align artistic styles with textual labels in a shared image-text embedding space. Alternatively, InST~\cite{zhang2023inversion} introduces style as a learnable textual embedding for each artwork image and learns the embedding through Textual Inversion training. 
The association of style with textual embeddings enables models to generalize to more fine-grained styles.

Improvements in style disentanglement techniques have powered a range of downstream applications including zero-shot generation, image-to-image translation, and art preservation through digital modeling. However, the abstract nature of style still poses a challenge
and understanding how neural networks internally represent and manipulate style requires further exploration.

\subsection{Diffusion-based Editing}

Diffusion-based models~\cite{rombach2022high, meng2021sdedit, kumari2023multi, ruiz2023dreambooth, gal2022image, han2023svdiff, nichol2021glide} demonstrate superior performance, excellent image quality, and decent computational efficiency for the tasks of image editing and generation. 
In general, diffusion models~\cite{ho2020denoising, song2021denoising, rombach2022high} gradually add noise to the image in the forward process, finally reaching a sample of the Gaussian distribution. In the reverse process, models generate samples by denoising from a random Gaussian noise sample. The Latent Diffusion Model (LDM)~\cite{rombach2022high} operates the forward/reverse processes in latent space rather than pixel space, significantly reducing computational time and cost. LDMs allow conditioning for text-to-image generation through the CLIP~\cite{radford2021learning} text encoder to process various prompts. Stable Diffusion (SD)~\cite{rombach2022high} is a popular image generation tool implemented as an LDM and can use diverse forms of conditions such as text, semantic maps, scratch sketches, and images for various applications, including Inpainting, Text-to-Image generation, Image-to-Image transformation, and Stylization tasks.

The need for higher fidelity and control in image generation models has led to the emergence of personalized diffusion-based techniques~\cite{ruiz2023dreambooth, gal2022image, kumari2023multi}. Personalization methods enable the fine-tuning of pretrained diffusion models using a small number of images, allowing for the incorporation of novel concepts, including specific objects, artistic styles, and facial biometrics. Techniques like Custom Diffusion~\cite{kumari2023multi}, Textual Inversion~\cite{gal2022image}, DreamBooth~\cite{ruiz2023dreambooth}, and Low-Rank Adaptations (LoRAs)~\cite{hu2022lora} represent different strategies for efficiently adapting diffusion models to personalized content. Figure~\ref{fig:inferenceVSfinetune} highlights the greater urgency of addressing style mimicry enabled by fine-tuning techniques.

\subsection{Perturbation-based Image Protection}



To prevent the misuse of images in diffusion-based editing and to weaken the effectiveness of such models, the seminal work~\cite{salman2023raising} proposed PhotoGuard which used perturbation-based protection.
Such protection methods~\cite{salman2023raising, van2023anti, liang2023adversarial, liang2023mist, shan2023glaze, shan2024nightshade, liu2024metacloak, chen2024editshield, mi2024visual, xu2024perturbing, pid, tang2025perturbation} can be broadly categorized along two dimensions: the type of image they are designed to protect and the stage of the generative process they target. 

In terms of image type, some methods focus specifically on artistic images, aiming to preserve the stylistic integrity and authorship of artworks. Methods such as Glaze~\cite{shan2023glaze}, Nightshade~\cite{shan2024nightshade}, and MIST~\cite{liang2023mist}, introduce imperceptible noise to prevent unauthorized style learning or mimicry. Other methods are designed to protect facial images and biometric data, such as Anti-DreamBooth~\cite{van2023anti}, MetaCloak~\cite{liu2024metacloak},  and VCPro~\cite{mi2024visual}, which disrupt identity extraction or facial reenactment in generative models.

In terms of interaction with the model, adversarial perturbation methods can be designed for protection against inference-based and finetuning-based editing. The methods of the former category~\cite{salman2023raising, chen2024editshield, shan2023glaze, pid} protect images from malicious exploiters to manipulate using a frozen model. Fine-tuning-time methods~\cite{van2023anti, liang2023adversarial, liu2024metacloak, xu2024perturbing} target the fine-tuning stage, poisoning the data to prevent the model from effectively learning sensitive information during adaptation. 
As fine-tuning gains traction as a preferred method for personalization and style transfer, training-time defenses have become increasingly critical for safeguarding identity, authorship, and stylistic integrity. Our work operates at the intersection of these concerns, aiming to protect artistic images from unauthorized style mimicry enabled by fine-tuning. 

\section{StyleProtect: Methodology}
\label{sec:methodology}

\subsection{Preliminary}

We describe the principles of general image generation using LDM and the personalization using fine-tuning techniques to provide a background for the editing scenario.

Suppose the image $x_0$ follows a real data distribution, $x_0 \sim q(x_0)$, the encoder of Autoencoder $\mathcal{E}_\phi$ first embeds $x_0$ into latent variable, $z_0 = \mathcal{E}_\phi(x_0)$, parameterized by $\phi$. The forward process keeps accumulating the noise during $T$ steps, following the Markov process. Note that $\beta$ schedules a sequence of noise values that determine how much noise is added at each step. 
Finally, $z_T$ becomes approximately an isotropic Gaussian random variable when $\overline{\alpha}_t \to 0$.
Meanwhile, the LDM allows a prompt $c$ as the guidance for the generation. The prompt $c$, embedded by the CLIP~\cite{radford2021learning} text encoder, is introduced into the cross-attention layers of the U-Net, achieving the mapping from text to images. The objective function of training LDMs is formulated as:
\begin{equation}
    \mathcal{L}_{LDM}(\theta, z_0) = \mathbb{E}_{z_0, t, \epsilon \sim \mathcal{N}(0,1),c} \| \epsilon-\epsilon_\theta (z_t,t,c)\|^2_2,
\label{eq: loss}
\end{equation}
where $\epsilon_\theta$ is the denoising network.

Personalization techniques~\cite{ruiz2023dreambooth, gal2022image, kumari2023multi} enable the adaptation of pretrained diffusion models to novel concepts using a small set of reference images. Given a small set of images depicting a specific subject, the model is trained to associate the visual appearance with a pseudo-token (e.g., [\textbf{V}] or [\textit{sks}]). During training, this token is embedded into the model’s learned representation space, allowing it to condition the generation process and reproduce the subject in varied textual prompts and scenes. To preserve the memory of existing tokens and maintain the model's inherent generative capabilities when learning new concepts, DreamBooth~\cite{ruiz2023dreambooth} incorporates a prior-preservation loss to prevent overfitting.
Equation~\ref{eq: loss_DB} defines the training objective, which consists of two terms: the standard denoising loss used in diffusion models (first term), and a regularization loss computed on class-generic samples with a prior prompt $c_{pr}$ (second term).
\begin{equation}
\begin{split}
\mathcal{L}_{DB}(\theta, x_0) =\ 
&\mathbb{E}_{x_0, t, t'} \left\| \epsilon - \epsilon_\theta (x_{t+1}, t, c) \right\|_2^2 \\
&+ \lambda \left\| \epsilon' - \epsilon_\theta(x'_{t'+1}, t', c_{pr}) \right\|_2^2
\end{split}
\label{eq: loss_DB}
\end{equation}
where $\epsilon,\epsilon'\sim \mathcal{N}(0,I)$.
$\lambda$ emphasizes the importance of the prior presentation loss term, and $x'_{t'+1}$ is the noisy variable of the class example $x'$ that is generated from the original diffusion model weights $\theta_{ori}$ with $c_{pr}$. DreamBooth serves as the foundational method for subject-driven fine-tuning. In contrast, later methods propose more parameter-efficient alternatives: Custom Diffusion~\cite{kumari2023multi} updates only cross-attention layers; Textual Inversion~\cite{gal2022image} optimizes learned text embeddings; and LoRA~\cite{hu2022lora} injects trainable low-rank matrices into frozen weights. We utilize DreamBooth in the mimicry scenario, as it is the most widely used and foundational approach for fully adapting diffusion models to new visual concepts.

\subsection{Style-based Cross-attention Layer Selection}

As mentioned, Custom Diffusion optimizes the parameters of cross-attention layers, rather than the whole model, significantly reducing computational overhead while maintaining high-quality generation performance. Based on this idea, CAAT~\cite{xu2024perturbing} further demonstrates that although cross-attention layers contain the fewest parameters, they play a critical role in model optimization. Inspired by these findings, we propose our method based on a hypothesis that within the cross-attention layers, certain layers may be particularly sensitive to artistic style representations.

\begin{figure}[ht]
    \centering
    \begin{subfigure}[b]{\linewidth}
        \centering
        \includegraphics[width=\linewidth]{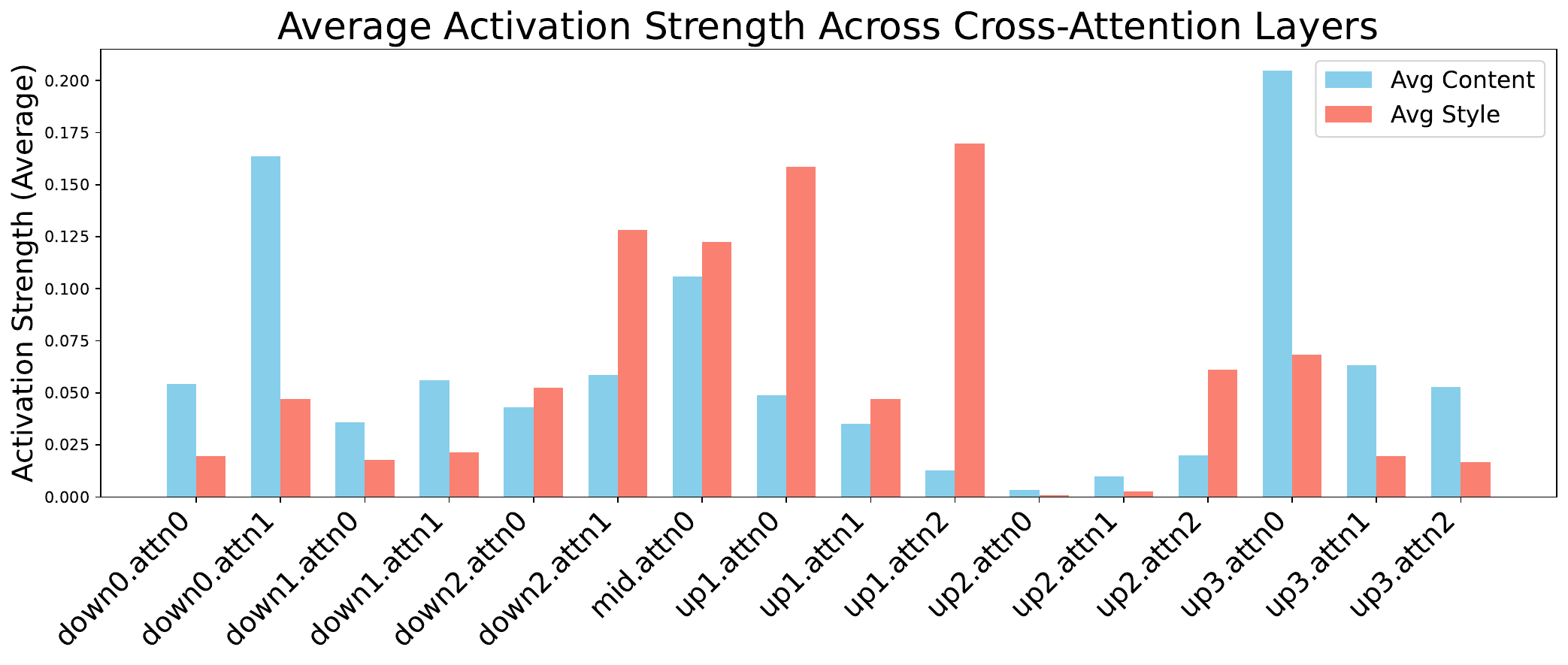}
        \caption{Mean activation strength associated with style and content tokens across cross-attention layers.}

        \label{fig: activation_map}
    \end{subfigure}

    \begin{subfigure}[b]{\linewidth}
        \centering
        \includegraphics[width=\linewidth]{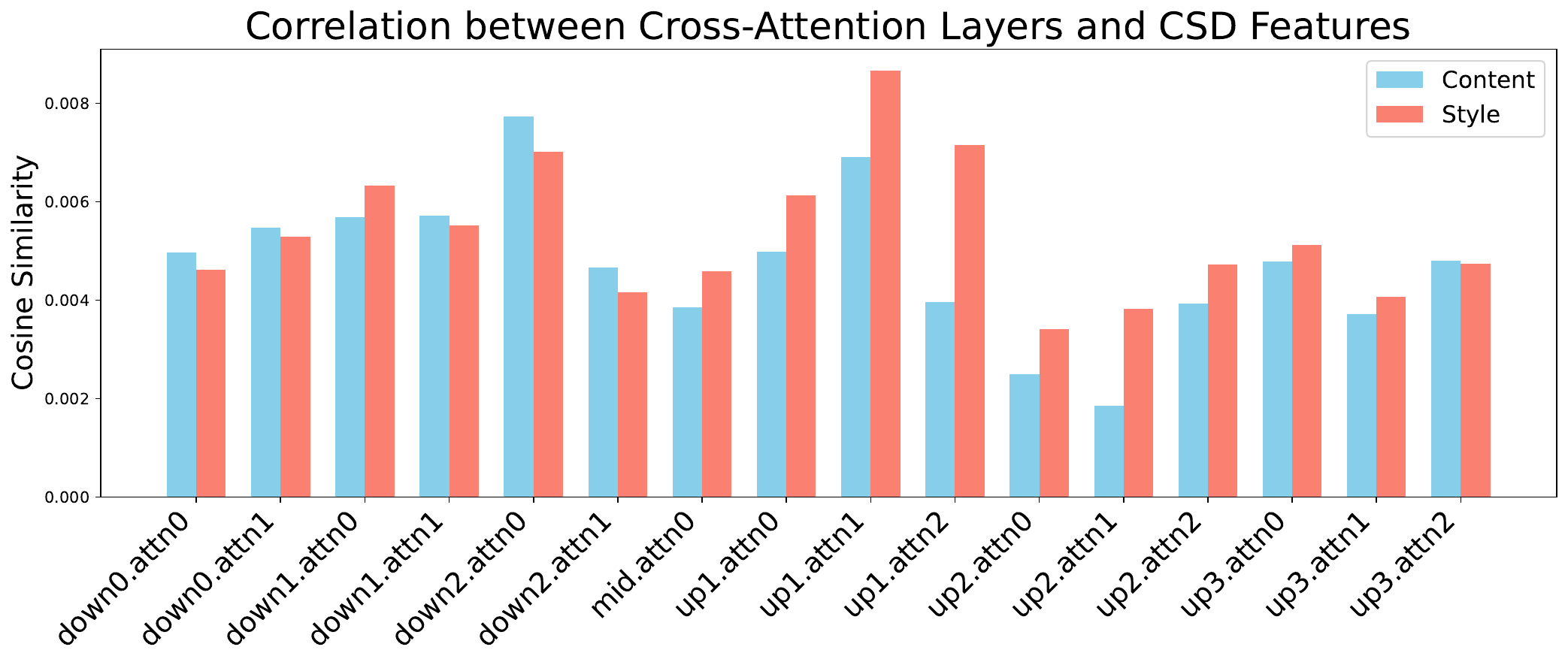}
        \caption{Cosine similarity between style-based descriptors and attention maps over layers. }
        \label{fig: activation_csd}
    \end{subfigure}

    \caption{The diagrams show a consistent trend: middle cross-attention layers exhibit stronger activation for style tokens than for content tokens, indicating heightened sensitivity to stylistic features. Abbreviations used: “down” refers to the down block, “mid” to the mid block, “up” to the up block, and “attn” to attention.}
    \label{fig:activation}
\end{figure}

To verify the hypothesis, we construct a set of diverse prompts that combine various content and artistic styles. Specifically, we define five common objects (e.g., dog, car, house) as content categories and select 30 well-known artists to represent distinct artistic styles. Each artist's name is treated as a proxy for their style. For example, the prompt ``a dog in the style of ClaudeMonet" pairs the content “dog” with the impressionist style of Monet.
These prompts are input into a text-to-image diffusion model~\cite{rombach2022high} to generate corresponding images. The U-Net in Stable Diffusion v1.5 contains 16 cross-attention layers — 6 in the down blocks, 1 in the middle block, and 9 in the up blocks. We extract attention maps from U-Net's cross-attention layers. By aggregating these values across all prompts, we visualize how different cross-attention layers respond to stylistic versus semantic features, in Figure.~\ref{fig: activation_map}. The figure illustrates and supports our hypothesis that certain cross-attention layers, particularly the middle layers, are more sensitive to style tokens.

For further evidence, we extract style descriptors using CSD~\cite{somepalli2024measuring} from the generated image and compute the cosine similarity between these features and each attention map. We provide details in the Supplemental Material. Figure~\ref{fig: activation_csd} illustrates the association between attention activation and style features extracted from an external model. Consistent with the activation strength patterns, the cosine similarity results indicate that mid-level (depth-wise) cross-attention layers predominantly encode stylistic information. Notably, ``up1.attn2" layer exhibits the greatest style vs. content divergence across both figures.

This two-fold analysis allows us to identify which attention layers are most involved in encoding style and how well their attention patterns align with the output visual style. By focusing optimization on these style-sensitive layers, we can model stylistic features more effectively. Conversely, to protect artworks, we can target these layers to update adversarial perturbations that weaken the model's ability to associate style-related visual and textual features, thereby disrupting style mimicry.

\subsection{Adversarial Perturbations}

\begin{figure}[!t]
    \centering
    \includegraphics[width=\linewidth]{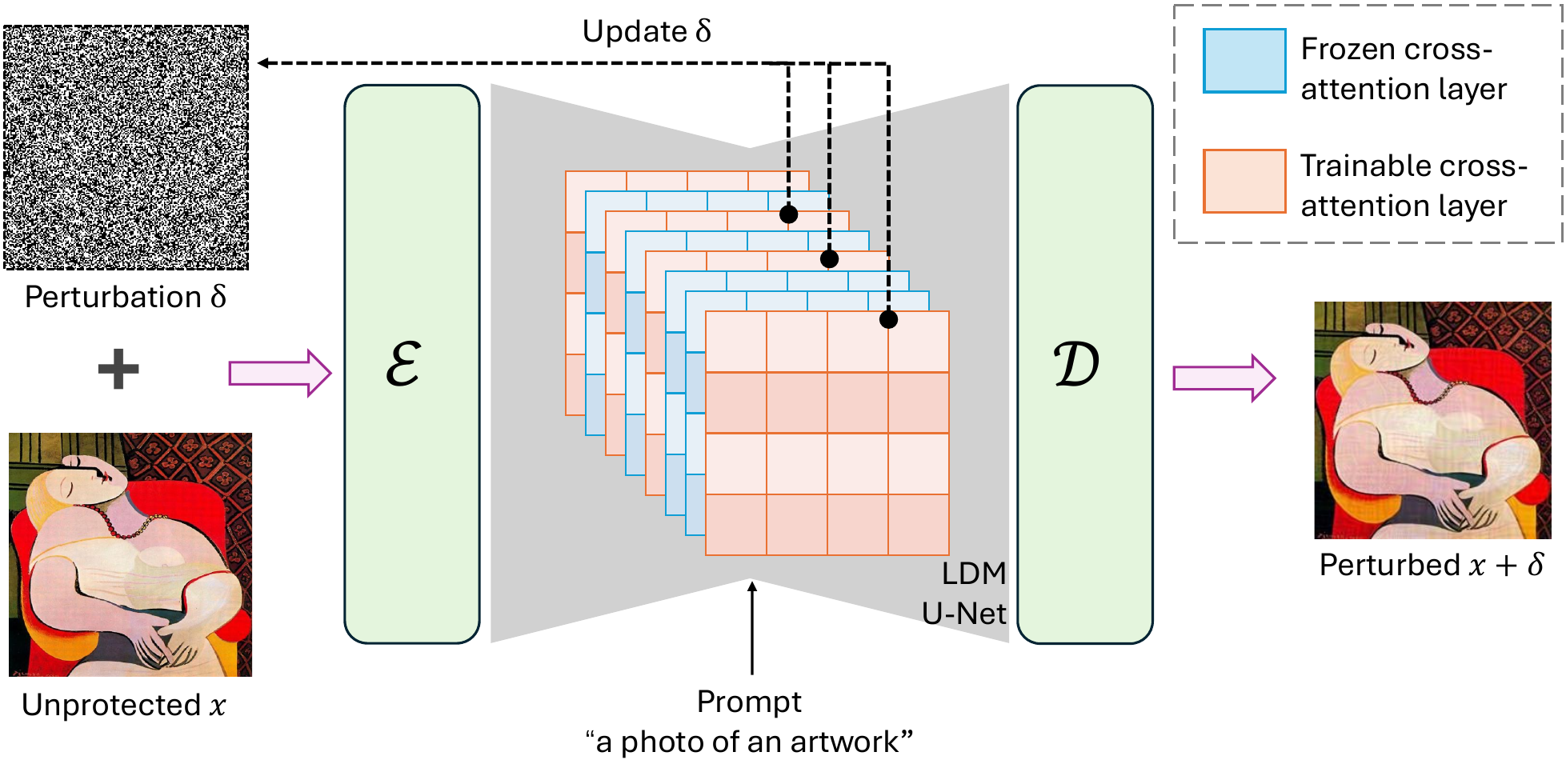}
    \caption{Overview of our pipeline for \textit{StyleProtect}. The proposed method updates the perturbation while fine-tuning only style relevant cross-attention layers. At each step, the selected layers are trained to enhance style representation, but the perturbation gradually causes the model to lose the aforementioned capabilities.}

    \label{fig:styleprotect}
\end{figure}

Based on the style vs. content tradeoff analysis of the cross-attention layers,
we propose \textit{StyleProtect}, a protective framework against style mimicry. 
In the framework, the majority of model parameters remain frozen, and we fine-tune selected cross-attention layers, following DreamBooth loss (see Eq.~\ref{eq: loss_DB}). We select the top four cross-attention layers with the highest style–content divergence, including three in the up blocks (up\_blocks.1/2) and one in the down block (down\_blocks.2). Specifically, the selected layers are:
up\_blocks.1.attentions.\{0,2\}, up\_blocks.2.attentions.2, and down\_blocks.2.attentions.1. Training of such layers strengthens style encoding. Concurrently, a perturbation $\delta$ is iteratively optimized on the input images to intentionally degrade the model’s ability to perceive and synthesize artistic styles. Figure~\ref{fig:styleprotect} shows the pipeline of our method. 

\section{Experiments}
\label{sec:experiments}

In this section, we describe the details of our experiments used to evaluate the performance of StyleProtect on fine-tuned diffusion models. 

\subsection{Datasets}

To construct a refined and style-representative artwork dataset, we begin with the WikiArt dataset~\cite{karras2019style}, which contains over 80,000 artworks spanning 129 artists, 11 genres, and 27 styles. Our goal is to curate a high-quality subset that captures distinctive artistic styles in a way that generative models can effectively recognize and learn from. We create a benchmark of 30 stylistically unique artists by leveraging ChatGPT-4~\cite{openai_chatgpt} to nominate candidates with distinct visual styles, manually verifying their work, and selecting 3-5 representative images per artist to evaluate style mimicry and protection. The details of the refined dataset construction are provided in the Supplemental Material. 
Besides, to evaluate the generalizability of the proposed approach and incorporate styles from the modern animation industry, we select five representative anime styles from the Anita dataset~\cite{Anita2024}. For each style, we construct a folder containing 4–5 keyframes extracted from animation videos. These keyframes are carefully chosen to consistently reflect the distinctive visual characteristics of each style, enabling reliable evaluation of style-specific protection performance.

\subsection{Experimental Settings}

We apply protection methods on clean images, including our proposed StyleProtect and several state-of-the-art methods~\cite{salman2023raising, liang2023mist, shan2023glaze, xu2024perturbing}. Then, we fine-tune diffusion models using subsets of both unprotected and protected images, corresponding to each distinct artist. During fine-tuning, the model learns to associate pseudo-labels with specific artistic styles. This enables us to generate images conditioned on content prompts while mimicking the artistic style. A more successful protection would produce fewer style elements of the original artist in the fine-tuned model's outputs.

\subsection{Metrics}
To evaluate the effectiveness of various protection mechanisms, we assess both the invisibility of perturbations and the fidelity of style replication. Excessive or conspicuous perturbations may alert attackers and undermine the artwork’s authenticity. Even slight noise can be noticeable in artworks, making imperceptibility a critical metric. We measure the difference between original (clean) and protected images using PSNR, SSIM, and LPIPS~\cite{zhang2018unreasonable}, which capture differences at structural, noise, and perceptual levels, respectively. Beyond visual fidelity, it is essential to assess whether style replication remains effective after applying protection. We evaluate style protection performance using style-based metrics, including ArtFID~\cite{wright2022artfid}, which computes the Fréchet distance between feature distributions extracted by a pretrained Inception network, and the CSD score~\cite{somepalli2024measuring}, which measures the cosine similarity between extracted style descriptors. These metrics allow us to compare the styles of edits of protected images with their original artworks, assessing the extent of style replication. We also report the runtime of each method per image for efficiency.

\subsection{Implementation Details} Our experiments were conducted on a workstation with four NVIDIA RTX A5000 GPUs (24 GB each). To ensure reproducibility, we fix the random seed to 9222, thereby guaranteeing consistent and repeatable results across different runs. Both the backbone model used in StyleProtect and the fine-tuned models are based on Stable Diffusion v1.5~\cite{rombach2022high}, while baseline methods are evaluated using their respective default settings. For performance evaluation of the fine-tuned models, we employ a set of content prompts designed to assess style replication capability. Examples of such prompts include: “A boy riding a bicycle on a sunny street,” “A dog sitting under a tree in a park,” and “An old lady reading a newspaper on a bench.” A full list of prompts will be released with the code and data for reproducibility~\footnote{https://anonymous.4open.science/r/StyleProtect}.

\section{Results}

\begin{table*}[!ht]
\def\arraystretch{1.2}
    \centering
        \begin{tabular}{l|ccc|cc|c}
        \toprule
        & \multicolumn{3}{c|}{Invisibility} & \multicolumn{2}{c|}{Style Protection} & \multicolumn{1}{c}{Efficiency}\\ 
        \cline{2-4} \cline{5-6} \cline{7-7}
        & PSNR $\uparrow$ & SSIM $\uparrow$ & LPIPS $\downarrow$  & ArtFID $\uparrow$ & CSD cos-sim $\downarrow$ & Runtime (s/img) $\downarrow$ \\ \hline
        Unprotected (clean) & 100 & 1 & 0 & 32.83 & 0.28 & - \\
        PhotoGuard~\cite{salman2023raising} & 29.58 & \underline{0.98} & \underline{0.13} & 29.62 & 0.47 & 102.84  \\ 
        Glaze~\cite{shan2023glaze} & \textbf{31.38} & \textbf{0.99} & \textbf{0.06} & 30.95 & 0.36 & 62.50  \\ 
        MIST~\cite{liang2023mist} & 26.35 & 0.97 & 0.26 & 31.15 & \underline{0.24} & 8280$^\dagger$\\ 
        Anti-DB~\cite{van2023anti} & \underline{29.96} & \underline{0.98} & 0.16 & \underline{31.33} & 0.32 & \textbf{33.34} \\
        CAAT~\cite{xu2024perturbing} & 25.81 & 0.80 & 0.25 & 30.30 & 0.33 & 53.62 \\ 
        \textit{StyleProtect} & 27.70 & \underline{0.98} & 0.26 & \textbf{32.97} & \textbf{0.23} & \underline{39.20} \\
        \bottomrule
    \end{tabular}
    \caption{Performance of different protection methods compared to an unprotected baseline, evaluated on invisibility, style protection and efficiency. The best performance values are shown in bold, and the second-best values are underlined. Our method demonstrates competitive imperceptibility, the second-best runtime and achieves the highest level of style protection. $^\dagger$Glaze is released as a software and was run on an Intel i7 CPU. All other runtimes are reported based on a RTX A5000 GPU.}
    \label{tab:results}
    
\end{table*}

We report quantitative results in Table.~\ref{tab:results} to demonstrate the performance of the proposed method compared with other state-of-the-art methods. 

\begin{figure*}[htb]
    \centering
    \includegraphics[width=\linewidth]{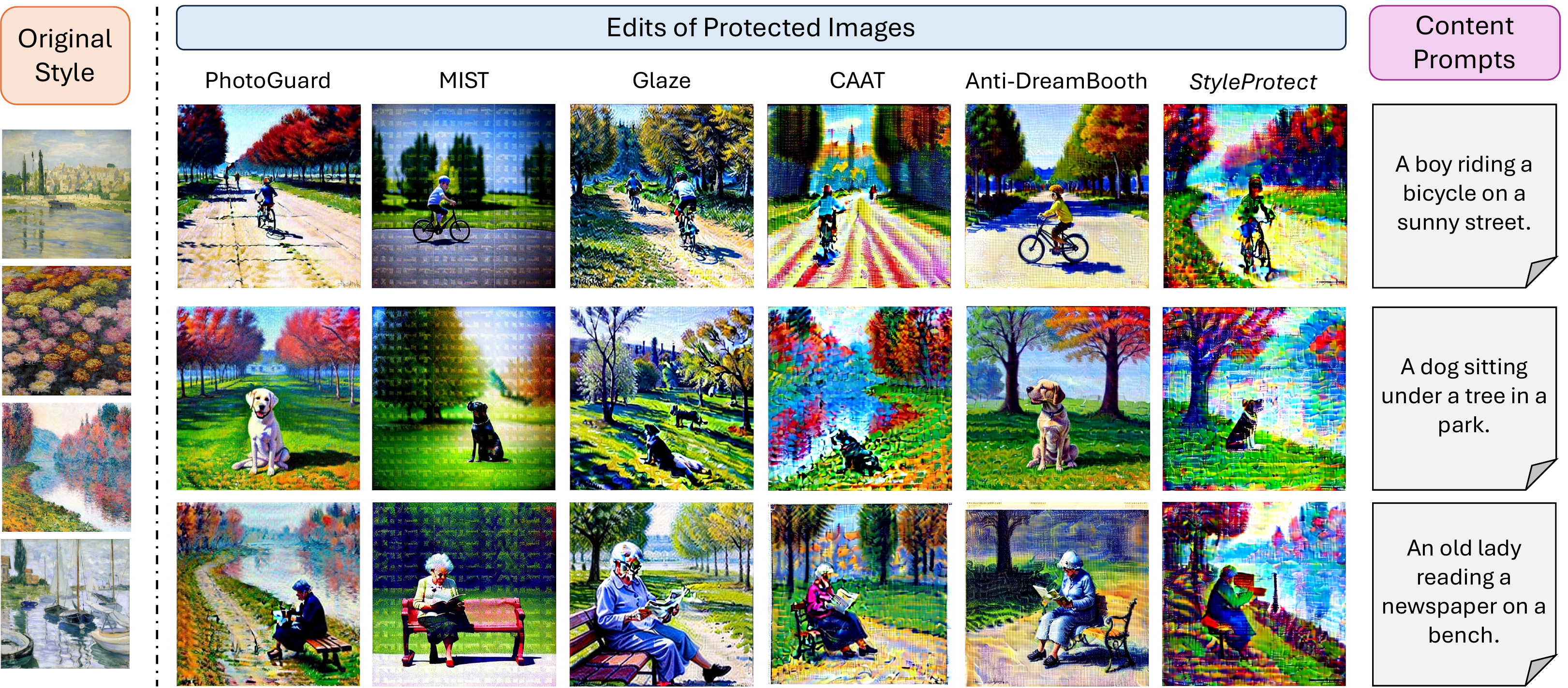}
    \caption{Visualization of edits aiming to replicate Claude Monet's style via fine-tuning techniques, under different protection methods. Each row corresponds to a distinct content prompt.}
    \label{fig:qualitative}
\end{figure*}

The imperceptibility of the perturbation is evaluated using PSNR, SSIM, and LPIPS metrics. Higher PSNR and SSIM values, along with lower LPIPS scores, indicate better visual similarity to the original images and greater invisibility of the perturbation. 
PSNR highlights differences in raw pixel values, whereas SSIM focuses on structural similarity, and LPIPS evaluates perceptual similarity by mimicking human vision. Notably, SSIM and LPIPS are more sensitive to structural and stylistic variations rather than absolute pixel differences. 
According to the result, our method demonstrates competitive imperceptibility. 
To assess the effectiveness of style replication, we employ style-based metrics, ArtFID and CSD. Higher ArtFID and lower CSD cosine similarity indicate weaker style fidelity, thus reflecting stronger protection effectiveness. Our proposed method, StyleProtect, achieves the highest performance on style protection metrics, demonstrating its effectiveness in preserving artistic style.
For efficiency performance, we calculate the runtime in seconds for protecting each image using the various methods. Our method ranks as the second fastest after Anti-DreamBooth which uses much fewer iterations.

In addition to quantitative evaluation, 
we visually compare edits produced by fine-tuned models with and without protective methods
in Figure.~\ref{fig:qualitative}. Given three distinct content prompts, we visualize the generated edits that attempt to replicate Claude Monet’s artistic style using the fine-tuning technique based on protected inputs. Edits protected by our method exhibit a consistent and more distorted pattern, making them easily recognizable and disrupting the mimicry attempt. We also provide the perturbations' stealthiness for different protection methods in the Supplementary Material.

\begin{figure}[t]
    \centering
    \includegraphics[width=0.9\linewidth]{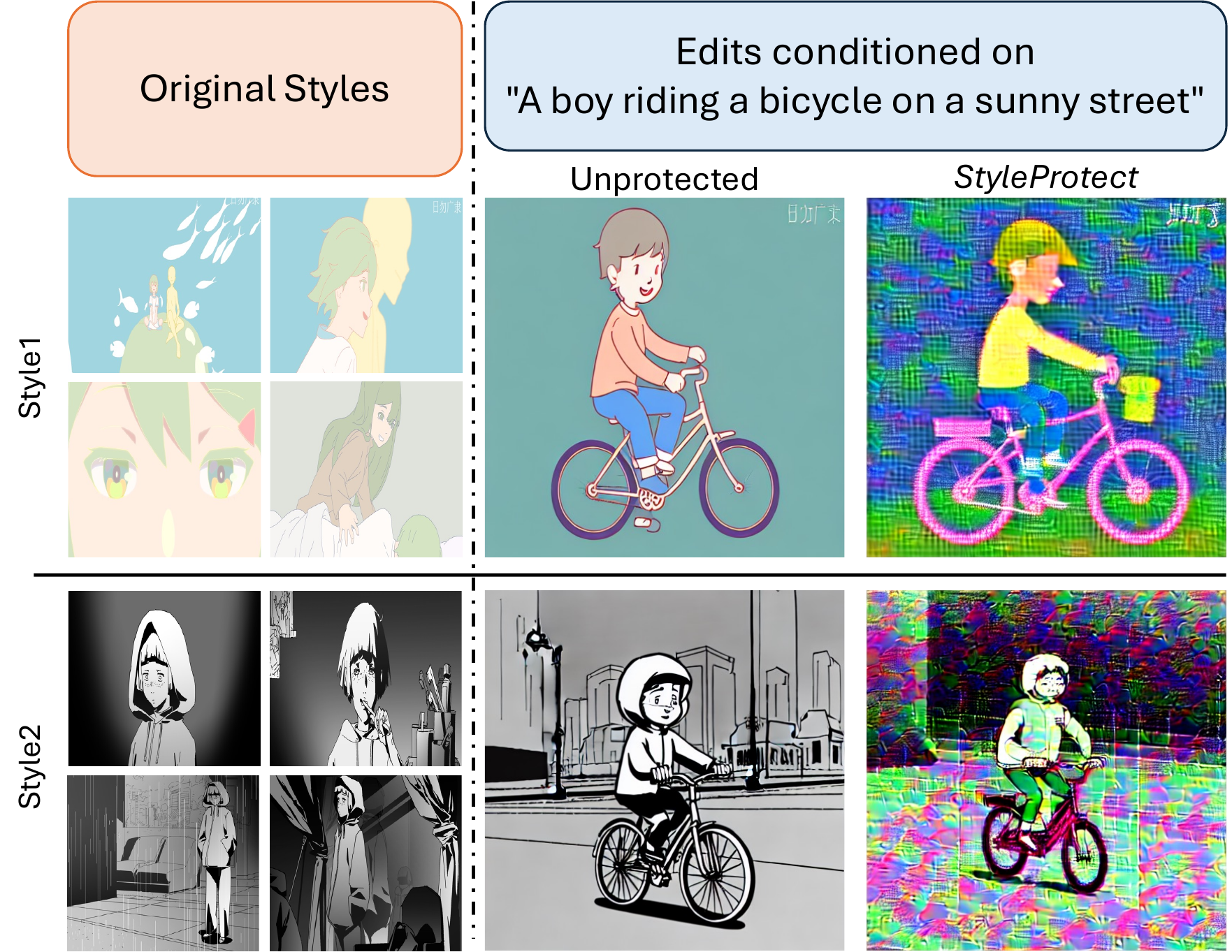}
    \caption{Visualization of edits replicating two cartoon styles from the Anita Dataset. We compare outputs from model fine-tuned on clean images versus those fine-tuned on images protected by our method, demonstrating that our approach effectively preserves cartoon styles.}
    \label{fig: anita}
\end{figure}

After demonstrating the effectiveness of our method on a well-known artwork dataset, we further evaluate its performance on a niche cartoon animation dataset. 
Table.~\ref{tab:quan_anita} presents the overall performance of StyleProtect in preventing style mimicry compared to unprotected images. 
To visualize the performance, we select two representative styles and compare the edited results of protected images using StyleProtect with those of unprotected images, as shown in Fig.~\ref{fig: anita}. 
Our method effectively protects cartoon styles both quantitatively and qualitatively.

\begin{table}[H]
\def\arraystretch{1.2}
 \centering
 \begin{tabular}{c|cc}
    \toprule
       & ArtFID $\uparrow$ & CSD $\downarrow$ \\
      \midrule
   Unprotected & 21.05 & 0.44 \\
   \textit{StyleProtect}  & \textbf{25.90} & \textbf{0.25} \\
   \bottomrule
 \end{tabular}
 \caption{Performance comparison of StyleProtect with unprotected images on Anita Dataset.}
 \label{tab:quan_anita}
\end{table}

\subsection{Ablation Studies}
\label{sec:ablationstudy}

\noindent\textbf{Impact of Different Layer Selection} - Our proposed method, StyleProtect, selects the top four layers with the highest style-content difference. To evaluate the significance of this layer selection strategy, we also evaluated two other variants. One selects the most style sensitive layer (up\_blocks.1.attentions.2) only while the other selects the middle five layers of the network that attend more to style than content (from down\_blocks.2.attentions.1 to up\_blocks.1.attentions.2). This is similar to the observations in~\cite{gatys2016image} where style was generally captured in mid-level image features layers of a CNN. The results show that our selection achieves the best performance based on invisibility and style protection. 

\begin{table}[htb]
\def\arraystretch{1.2}
\setlength{\tabcolsep}{1mm} 
 \centering
 \small
 \begin{tabular}{p{1.7cm}|p{1.1cm}p{1.0cm}p{1.1cm}|p{1.2cm}p{0.9cm}}
    \toprule
     & PSNR $\uparrow$ & SSIM $\uparrow$ & LPIPS $\downarrow$ & ArtFID $\uparrow$ & CSD $\downarrow$ \\
      \midrule
    Top sensitive & \underline{27.68} & \textbf{0.98} & \underline{0.27} & \textbf{33.30} & 0.25  \\
    5 Mid-level  & \underline{27.68} & \textbf{0.98} & \underline{0.27}  &  32.53 & \underline{0.24}  \\
    \textit{StyleProtect} & \textbf{27.70} & \textbf{0.98} & \textbf{0.26}  & \underline{32.97} & \textbf{0.23} \\
   \bottomrule
 \end{tabular}%
 \caption{Performance comparison for different style-based layer selections on invisibility metrics, ArtFID and CSD cosine similarity.}
 \label{tab:abla_layer}
\end{table}

\noindent\textbf{Impact of Perturbation Budget} - For adversarial perturbation-based protection~\cite{threat}, the budget is the maximum amount of change (perturbation) allowed to be applied to the input image to degrade the model's generation performance. It controls the strength and perceptibility of the adversarial modifications. To ablate the effect of perturbation strength, we evaluate our method under different budget levels, varying $\delta$ from 0.05 to 0.13. As shown in Tab.~\ref{tab:abla_budget}, StyleProtect offers an effective trade-off between stealthiness and style protection.



\begin{table}[htb]
\def\arraystretch{1.2}
 \centering
\small
 \begin{tabular}{c|ccc|cc}
    \toprule
     $\delta$ & PSNR $\uparrow$ & SSIM $\uparrow$ & LPIPS $\downarrow$ & ArtFID $\uparrow$ & CSD $\downarrow$ \\
      \midrule
   $0.1$ & 27.70 & \underline{0.98} & 0.26  & \textbf{32.97} & \underline{0.23} \\
   $0.05$  & \textbf{30.53} & \textbf{0.99} & \textbf{0.14} & 31.17 & 0.32 \\
   $0.08$  & \underline{28.73} & \underline{0.98} & \underline{0.22} & 31.33 & 0.28\\
   $0.13$  & 26.31 & 0.97 & 0.32 & \underline{32.34} & \textbf{0.22} \\
   \bottomrule
 \end{tabular}
 \caption{Performance comparison of StyleProtect (default $\delta=0.1$) with different perturbation budgets on invisibility and style comparison metrics. Lower budget improves imperceptibility but isn't as effective for protection.}
 \label{tab:abla_budget}
\end{table}

\noindent\textbf{Robustness Against Post-Processing Transformations} - 
Perturbation-based protection methods can be fragile and vulnerable to some post-processing countermeasures. To further assess the robustness of our method, we apply a series of simple image transformation operations, including Gaussian blurring using a kernel size of 3x3 with $\sigma$ set to 0.05, and JPEG compression with quality factor 75, to protected data and repeat our experiments. Our method demonstrates competitive performance against post-processing transformations, comparable to the strong protection offered by MIST, despite using a smaller perturbation budget.

\begin{table}[htb]
\small  
\def\arraystretch{1.2}
\centering
    \begin{tabular}{l cc||cc}
    \toprule
     & \multicolumn{2}{c}{\textbf{w/ JPEG}} & \multicolumn{2}{c}{\textbf{w/ Gaussian blur}} \\
    \cline{2-3} \cline{4-5}
    Method & ArtFID $\uparrow$ & CSD $\downarrow$ & ArtFID $\uparrow$ & CSD $\downarrow$ \\
    \midrule
    \textit{StyleProtect}    & \textbf{32.29} & \underline{0.26} & \underline{32.25} & \underline{0.25} \\
    Anti-DB & 30.73 & 0.32 & 31.39 & 0.31 \\
    CAAT            & 30.83 & 0.32 & 30.90 & 0.33 \\
    MIST            & \underline{31.89} & \textbf{0.20} & \textbf{32.29} & \textbf{0.22} \\
    Glaze           & 31.32 & 0.40 & 30.27 & 0.38 \\
    PhotoGuard      & 29.10 & 0.42 & 29.28 & 0.44 \\
    \bottomrule
    \end{tabular}%
\caption{Performance comparison of robustness on ArtFID and CSD cosine similarity.}
\label{tab:performance_metrics}
\end{table}

\section{Conclusion}
\label{sec:conclusion}

In this paper, we investigate the sensitivity of cross-attention layers in diffusion models to artistic style representations. Based on this observation, we propose StyleProtect, a lightweight protection mechanism that optimizes imperceptible perturbations to undermine style preservation. Our approach effectively disrupts unauthorized style replication while maintaining the original image quality. Extensive experiments on the refined WikiArt and cartoon Anita datasets demonstrate that StyleProtect achieves strong protection while maintaining stealthiness. 

\noindent\textbf{Limitations and future direction} This work primarily focuses on stylistic protection against Stable Diffusion v1.5. In future research, we plan to assess the generalizability of our approach across different generative architectures. 
%
Although, there has been progress in learning style as a concept, 
there is scope for improvement in style representation and metrics that can be used to measure style protection.

\bibliography{aaai2026}

\begin{thebibliography}{4}
\providecommand{\natexlab}[1]{#1}

\bibitem[{Karras, Laine, and Aila(2019)}]{karras2019style}
Karras, T.; Laine, S.; and Aila, T. 2019.
\newblock A style-based generator architecture for generative adversarial networks.
\newblock In \emph{Proceedings of the IEEE/CVF conference on computer vision and pattern recognition (CVPR)}, 4401--4410.

\bibitem[{OpenAI(2025)}]{openai_chatgpt}
OpenAI. 2025.
\newblock ChatGPT.
\newblock \url{https://chat.openai.com/}.
\newblock Accessed: 2025-01-10.

\bibitem[{Somepalli et~al.(2024)Somepalli, Gupta, Gupta, Palta, Goldblum, Geiping, Shrivastava, and Goldstein}]{somepalli2024measuring}
Somepalli, G.; Gupta, A.; Gupta, K.; Palta, S.; Goldblum, M.; Geiping, J.; Shrivastava, A.; and Goldstein, T. 2024.
\newblock Measuring Style Similarity in Diffusion Models.
\newblock \emph{Proceedings of the European conference on computer vision (ECCV)}.

\bibitem[{Tan et~al.(2019)Tan, Chan, Aguirre, and Tanaka}]{artgan2018}
Tan, W.~R.; Chan, C.~S.; Aguirre, H.; and Tanaka, K. 2019.
\newblock Improved ArtGAN for Conditional Synthesis of Natural Image and Artwork.
\newblock \emph{IEEE Transactions on Image Processing}, 28(1): 394--409.

\end{thebibliography}


\begin{thebibliography}{62}
\providecommand{\natexlab}[1]{#1}

\bibitem[{Ahn et~al.(2024)Ahn, Lee, Lee, Kim, Kim, Nam, and Hong}]{ahn2024dreamstyler}
Ahn, N.; Lee, J.; Lee, C.; Kim, K.; Kim, D.; Nam, S.-H.; and Hong, K. 2024.
\newblock Dreamstyler: Paint by style inversion with text-to-image diffusion models.
\newblock In \emph{Proceedings of the AAAI Conference on Artificial Intelligence}, volume~38, 674--681.

\bibitem[{Ahn et~al.(2025)Ahn, Yoo, Ahn, Kim, and Nam}]{ahn2024nearly}
Ahn, N.; Yoo, K.; Ahn, W.; Kim, D.; and Nam, S.-H. 2025.
\newblock Nearly Zero-Cost Protection Against Mimicry by Personalized Diffusion Models.
\newblock In \emph{Proceedings of the IEEE/CVF Conference on Computer Vision and Pattern Recognition (CVPR)}.

\bibitem[{Akhtar and Mian(2018)}]{threat}
Akhtar, N.; and Mian, A. 2018.
\newblock Threat of Adversarial Attacks on Deep Learning in Computer Vision: A Survey.
\newblock \emph{IEEE Access}, 6: 14410--14430.

\bibitem[{Baio(2022)}]{jackson2022invasive}
Baio, A. 2022.
\newblock Invasive Diffusion: How One Unwilling Illustrator Found Herself Turned into an AI Model.

\bibitem[{Chen et~al.(2024)Chen, Jin, Liu, Chen, Wang, and Sun}]{chen2024editshield}
Chen, R.; Jin, H.; Liu, Y.; Chen, J.; Wang, H.; and Sun, L. 2024.
\newblock Editshield: Protecting unauthorized image editing by instruction-guided diffusion models.
\newblock In \emph{European Conference on Computer Vision}, 126--142. Springer.

\bibitem[{Chu and Wu(2018)}]{chu2018image}
Chu, W.-T.; and Wu, Y.-L. 2018.
\newblock Image style classification based on learnt deep correlation features.
\newblock \emph{IEEE Transactions on Multimedia}, 20(9): 2491--2502.

\bibitem[{Cui et~al.(2025{\natexlab{a}})Cui, Ren, Lin, Xu, He, Xing, Lyu, Fan, Liu, and Tang}]{cui2025ft}
Cui, Y.; Ren, J.; Lin, Y.; Xu, H.; He, P.; Xing, Y.; Lyu, L.; Fan, W.; Liu, H.; and Tang, J. 2025{\natexlab{a}}.
\newblock Ft-shield: A watermark against unauthorized fine-tuning in text-to-image diffusion models.
\newblock \emph{ACM SIGKDD Explorations Newsletter}, 26(2): 76--88.

\bibitem[{Cui et~al.(2025{\natexlab{b}})Cui, Ren, Xu, He, Liu, Sun, Xing, and Tang}]{cui2025diffusionshield}
Cui, Y.; Ren, J.; Xu, H.; He, P.; Liu, H.; Sun, L.; Xing, Y.; and Tang, J. 2025{\natexlab{b}}.
\newblock DiffusionShield: A Watermark for Data Copyright Protection against Generative Diffusion Models.
\newblock \emph{SIGKDD Explor. Newsl.}, 26(2): 60–75.

\bibitem[{Gal et~al.(2023)Gal, Alaluf, Atzmon, Patashnik, Bermano, Chechik, and Cohen-or}]{gal2022image}
Gal, R.; Alaluf, Y.; Atzmon, Y.; Patashnik, O.; Bermano, A.~H.; Chechik, G.; and Cohen-or, D. 2023.
\newblock An Image is Worth One Word: Personalizing Text-to-Image Generation using Textual Inversion.
\newblock In \emph{The Eleventh International Conference on Learning Representations (ICLR)}.

\bibitem[{Gatys, Ecker, and Bethge(2016)}]{gatys2016image}
Gatys, L.~A.; Ecker, A.~S.; and Bethge, M. 2016.
\newblock Image style transfer using convolutional neural networks.
\newblock In \emph{Proceedings of the IEEE conference on computer vision and pattern recognition}, 2414--2423.

\bibitem[{Gibson(1966)}]{gibson1966senses}
Gibson, J.~J. 1966.
\newblock \emph{The Senses Considered as Perceptual Systems}.
\newblock Boston: Houghton Mifflin.

\bibitem[{Goodfellow et~al.(2014)Goodfellow, Pouget-Abadie, Mirza, Xu, Warde-Farley, Ozair, Courville, and Bengio}]{goodfellow2014generative}
Goodfellow, I.~J.; Pouget-Abadie, J.; Mirza, M.; Xu, B.; Warde-Farley, D.; Ozair, S.; Courville, A.; and Bengio, Y. 2014.
\newblock Generative adversarial nets.
\newblock \emph{Advances in neural information processing systems}, 27.

\bibitem[{Han et~al.(2023)Han, Li, Zhang, Milanfar, Metaxas, and Yang}]{han2023svdiff}
Han, L.; Li, Y.; Zhang, H.; Milanfar, P.; Metaxas, D.; and Yang, F. 2023.
\newblock Svdiff: Compact parameter space for diffusion fine-tuning.
\newblock In \emph{Proceedings of the IEEE/CVF International Conference on Computer Vision}, 7323--7334.

\bibitem[{Ho, Jain, and Abbeel(2020)}]{ho2020denoising}
Ho, J.; Jain, A.; and Abbeel, P. 2020.
\newblock Denoising diffusion probabilistic models.
\newblock \emph{Advances in neural information processing systems}, 33: 6840--6851.

\bibitem[{Hu et~al.(2022)Hu, Shen, Wallis, Allen-Zhu, Li, Wang, Wang, and Chen}]{hu2022lora}
Hu, E.~J.; Shen, Y.; Wallis, P.; Allen-Zhu, Z.; Li, Y.; Wang, S.; Wang, L.; and Chen, W. 2022.
\newblock Lo{RA}: Low-Rank Adaptation of Large Language Models.
\newblock In \emph{International Conference on Learning Representations (ICLR)}.

\bibitem[{Hu, Zhuang, and Gao(2024)}]{hu2024diffusest}
Hu, Y.; Zhuang, C.; and Gao, P. 2024.
\newblock DiffuseST: Unleashing the Capability of the Diffusion Model for Style Transfer.
\newblock In \emph{Proceedings of the 6th ACM International Conference on Multimedia in Asia}, 1--1.

\bibitem[{Huang et~al.(2018)Huang, Liu, Belongie, and Kautz}]{Huang_2018_ECCV}
Huang, X.; Liu, M.-Y.; Belongie, S.; and Kautz, J. 2018.
\newblock Multimodal Unsupervised Image-to-image Translation.
\newblock In \emph{Proceedings of the European Conference on Computer Vision (ECCV)}.

\bibitem[{Jing et~al.(2019)Jing, Yang, Feng, Ye, Yu, and Song}]{jing2019neural}
Jing, Y.; Yang, Y.; Feng, Z.; Ye, J.; Yu, Y.; and Song, M. 2019.
\newblock Neural style transfer: A review.
\newblock \emph{IEEE transactions on visualization and computer graphics}, 26(11): 3365--3385.

\bibitem[{Karras, Laine, and Aila(2019)}]{karras2019style}
Karras, T.; Laine, S.; and Aila, T. 2019.
\newblock A style-based generator architecture for generative adversarial networks.
\newblock In \emph{Proceedings of the IEEE/CVF conference on computer vision and pattern recognition (CVPR)}, 4401--4410.

\bibitem[{{Kevin Roose}(2022)}]{nyt_ai_art_2022}
{Kevin Roose}. 2022.
\newblock An A.I.-Generated Picture Won an Art Prize. Artists Aren’t Happy.

\bibitem[{Kumari et~al.(2023)Kumari, Zhang, Zhang, Shechtman, and Zhu}]{kumari2023multi}
Kumari, N.; Zhang, B.; Zhang, R.; Shechtman, E.; and Zhu, J.-Y. 2023.
\newblock Multi-concept customization of text-to-image diffusion.
\newblock In \emph{IEEE/CVF Conference on Computer Vision and Pattern Recognition (CVPR)}, 1931--1941.

\bibitem[{Li et~al.(2024{\natexlab{a}})Li, Mo, Li, and Wang}]{pid}
Li, A.; Mo, Y.; Li, M.; and Wang, Y. 2024{\natexlab{a}}.
\newblock PID: Prompt-Independent Data Protection Against Latent Diffusion Models.
\newblock In \emph{Proceedings of the 41st International Conference on Machine Learning (ICML)}.

\bibitem[{Li et~al.(2024{\natexlab{b}})Li, Tian, Li, Deng, and He}]{li2024autoregressive}
Li, T.; Tian, Y.; Li, H.; Deng, M.; and He, K. 2024{\natexlab{b}}.
\newblock Autoregressive image generation without vector quantization.
\newblock \emph{Advances in Neural Information Processing Systems}, 37: 56424--56445.

\bibitem[{Li et~al.(2017)Li, Wang, Liu, and Hou}]{li2017demystifying}
Li, Y.; Wang, N.; Liu, J.; and Hou, X. 2017.
\newblock Demystifying neural style transfer.
\newblock \emph{International Joint Conference on Artificial Intelligence (IJCAI)}.

\bibitem[{Liang and Wu(2023)}]{liang2023mist}
Liang, C.; and Wu, X. 2023.
\newblock Mist: Towards improved adversarial examples for diffusion models.
\newblock arXiv:2305.12683.

\bibitem[{Liang et~al.(2023)Liang, Wu, Hua, Zhang, Xue, Song, Xue, Ma, and Guan}]{liang2023adversarial}
Liang, C.; Wu, X.; Hua, Y.; Zhang, J.; Xue, Y.; Song, T.; Xue, Z.; Ma, R.; and Guan, H. 2023.
\newblock Adversarial example does good: Preventing painting imitation from diffusion models via adversarial examples.
\newblock In \emph{International Conference on Machine Learning (ICML)}, 20763--20786. PMLR.

\bibitem[{Liu et~al.(2025)Liu, Shah, Cui, and Lazebnik}]{liu2024unziplora}
Liu, C.; Shah, V.; Cui, A.; and Lazebnik, S. 2025.
\newblock Unziplora: Separating content and style from a single image.
\newblock \emph{IEEE/CVF International Conference on Computer Vision (ICCV)}.

\bibitem[{Liu et~al.(2024)Liu, Fan, Dai, Chen, Zhou, and Sun}]{liu2024metacloak}
Liu, Y.; Fan, C.; Dai, Y.; Chen, X.; Zhou, P.; and Sun, L. 2024.
\newblock MetaCloak: Preventing Unauthorized Subject-driven Text-to-image Diffusion-based Synthesis via Meta-learning.
\newblock In \emph{IEEE/CVF Conference on Computer Vision and Pattern Recognition (CVPR)}, 24219--24228.

\bibitem[{Lun, Kalogerakis, and Sheffer(2015)}]{lun2015elements}
Lun, Z.; Kalogerakis, E.; and Sheffer, A. 2015.
\newblock Elements of style: learning perceptual shape style similarity.
\newblock \emph{ACM Transactions on graphics (TOG)}, 34(4): 1--14.

\bibitem[{Matsuo and Yanai(2016)}]{matsuo2016cnn}
Matsuo, S.; and Yanai, K. 2016.
\newblock CNN-based style vector for style image retrieval.
\newblock In \emph{Proceedings of the 2016 ACM on International Conference on Multimedia Retrieval}, 309--312.

\bibitem[{Meng et~al.(2022)Meng, He, Song, Song, Wu, Zhu, and Ermon}]{meng2021sdedit}
Meng, C.; He, Y.; Song, Y.; Song, J.; Wu, J.; Zhu, J.-Y.; and Ermon, S. 2022.
\newblock {SDE}dit: Guided Image Synthesis and Editing with Stochastic Differential Equations.
\newblock In \emph{International Conference on Learning Representations (ICLR)}.

\bibitem[{Mi et~al.(2024)Mi, Tang, Cao, Li, and Liu}]{mi2024visual}
Mi, X.; Tang, F.; Cao, J.; Li, P.; and Liu, Y. 2024.
\newblock Visual-friendly concept protection via selective adversarial perturbations.
\newblock arXiv:2408.08518.

\bibitem[{{Natalie Proulx}(2022)}]{nyt_ai_generated_art_2022}
{Natalie Proulx}. 2022.
\newblock Are A.I.-Generated Pictures Art?

\bibitem[{Nichol et~al.(2021)Nichol, Dhariwal, Ramesh, Shyam, Mishkin, McGrew, Sutskever, and Chen}]{nichol2021glide}
Nichol, A.; Dhariwal, P.; Ramesh, A.; Shyam, P.; Mishkin, P.; McGrew, B.; Sutskever, I.; and Chen, M. 2021.
\newblock Glide: Towards photorealistic image generation and editing with text-guided diffusion models.
\newblock arXiv:2112.10741.

\bibitem[{OpenAI(2025)}]{openai_chatgpt}
OpenAI. 2025.
\newblock ChatGPT.
\newblock \url{https://chat.openai.com/}.
\newblock Accessed: 2025-01-10.

\bibitem[{Pan and Zhu(2024)}]{Anita2024}
Pan, Z.; and Zhu, Y. 2024.
\newblock Anita Dataset.
\newblock \url{https://zhenglinpan.github.io/AnitaDataset_homepage/}.
\newblock Accessed: 2024-06-24.

\bibitem[{Patashnik et~al.(2021)Patashnik, Wu, Shechtman, Cohen-Or, and Lischinski}]{patashnik2021styleclip}
Patashnik, O.; Wu, Z.; Shechtman, E.; Cohen-Or, D.; and Lischinski, D. 2021.
\newblock Styleclip: Text-driven manipulation of stylegan imagery.
\newblock In \emph{Proceedings of the IEEE/CVF international conference on computer vision (ICCV)}, 2085--2094.

\bibitem[{Radford et~al.(2021)Radford, Kim, Hallacy, Ramesh, Goh, Agarwal, Sastry, Askell, Mishkin, Clark et~al.}]{radford2021learning}
Radford, A.; Kim, J.~W.; Hallacy, C.; Ramesh, A.; Goh, G.; Agarwal, S.; Sastry, G.; Askell, A.; Mishkin, P.; Clark, J.; et~al. 2021.
\newblock Learning transferable visual models from natural language supervision.
\newblock In \emph{International conference on machine learning (ICML)}, 8748--8763. PMLR.

\bibitem[{Radford, Metz, and Chintala(2016)}]{radford2015unsupervised}
Radford, A.; Metz, L.; and Chintala, S. 2016.
\newblock Unsupervised Representation Learning with Deep Convolutional Generative Adversarial Networks.
\newblock In Bengio, Y.; and LeCun, Y., eds., \emph{4th International Conference on Learning Representations, {(ICLR)} 2016, San Juan, Puerto Rico, May 2-4, 2016, Conference Track Proceedings}.

\bibitem[{Richardson et~al.(2021)Richardson, Alaluf, Patashnik, Nitzan, Azar, Shapiro, and Cohen-Or}]{Richardson_2021_CVPR}
Richardson, E.; Alaluf, Y.; Patashnik, O.; Nitzan, Y.; Azar, Y.; Shapiro, S.; and Cohen-Or, D. 2021.
\newblock Encoding in Style: A StyleGAN Encoder for Image-to-Image Translation.
\newblock In \emph{Proceedings of the IEEE/CVF Conference on Computer Vision and Pattern Recognition (CVPR)}, 2287--2296.

\bibitem[{Rombach et~al.(2021)Rombach, Blattmann, Lorenz, Esser, and Ommer}]{rombach2022high}
Rombach, R.; Blattmann, A.; Lorenz, D.; Esser, P.; and Ommer, B. 2021.
\newblock High-resolution image synthesis with latent diffusion models. 2022 IEEE.
\newblock In \emph{CVF Conference on Computer Vision and Pattern Recognition (CVPR)}, volume~1.

\bibitem[{Ruiz et~al.(2023)Ruiz, Li, Jampani, Pritch, Rubinstein, and Aberman}]{ruiz2023dreambooth}
Ruiz, N.; Li, Y.; Jampani, V.; Pritch, Y.; Rubinstein, M.; and Aberman, K. 2023.
\newblock Dreambooth: Fine tuning text-to-image diffusion models for subject-driven generation.
\newblock In \emph{Proceedings of the IEEE/CVF Conference on Computer Vision and Pattern Recognition (CVPR)}, 22500--22510.

\bibitem[{Ruta et~al.(2021)Ruta, Motiian, Faieta, Lin, Jin, Filipkowski, Gilbert, and Collomosse}]{ruta2021aladin}
Ruta, D.; Motiian, S.; Faieta, B.; Lin, Z.; Jin, H.; Filipkowski, A.; Gilbert, A.; and Collomosse, J. 2021.
\newblock ALADIN: All layer adaptive instance normalization for fine-grained style similarity.
\newblock In \emph{Proceedings of the IEEE/CVF International Conference on Computer Vision (CVPR)}, 11926--11935.

\bibitem[{Sablatnig, Kammerer, and Zolda(1998)}]{sablatnig1998hierarchical}
Sablatnig, R.; Kammerer, P.; and Zolda, E. 1998.
\newblock Hierarchical classification of paintings using face-and brush stroke models.
\newblock In \emph{Proceedings. Fourteenth International Conference on Pattern Recognition}, volume~1, 172--174. IEEE.

\bibitem[{Salman et~al.(2023)Salman, Khaddaj, Leclerc, Ilyas, and Madry}]{salman2023raising}
Salman, H.; Khaddaj, A.; Leclerc, G.; Ilyas, A.; and Madry, A. 2023.
\newblock Raising the cost of malicious ai-powered image editing.
\newblock arXiv:2302.06588.

\bibitem[{Shan et~al.(2023)Shan, Cryan, Wenger, Zheng, Hanocka, and Zhao}]{shan2023glaze}
Shan, S.; Cryan, J.; Wenger, E.; Zheng, H.; Hanocka, R.; and Zhao, B.~Y. 2023.
\newblock Glaze: Protecting artists from style mimicry by Text-to-Image models.
\newblock In \emph{32nd USENIX Security Symposium (USENIX Security 23)}, 2187--2204.

\bibitem[{Shan et~al.(2024)Shan, Ding, Passananti, Wu, Zheng, and Zhao}]{shan2024nightshade}
Shan, S.; Ding, W.; Passananti, J.; Wu, S.; Zheng, H.; and Zhao, B.~Y. 2024.
\newblock Nightshade: Prompt-Specific Poisoning Attacks on Text-to-Image Generative Models.
\newblock In \emph{IEEE Symposium on Security and Privacy (S\&P)}.
\newblock ArXiv:2310.13828 [cs.CR].

\bibitem[{Silva et~al.(2021)Silva, Pratas, Antunes, Matos, and Pinho}]{silva2021automatic}
Silva, J.~M.; Pratas, D.; Antunes, R.; Matos, S.; and Pinho, A.~J. 2021.
\newblock Automatic analysis of artistic paintings using information-based measures.
\newblock \emph{Pattern Recognition}, 114: 107864.

\bibitem[{Somepalli et~al.(2024)Somepalli, Gupta, Gupta, Palta, Goldblum, Geiping, Shrivastava, and Goldstein}]{somepalli2024measuring}
Somepalli, G.; Gupta, A.; Gupta, K.; Palta, S.; Goldblum, M.; Geiping, J.; Shrivastava, A.; and Goldstein, T. 2024.
\newblock Measuring Style Similarity in Diffusion Models.
\newblock \emph{Proceedings of the European conference on computer vision (ECCV)}.

\bibitem[{Song, Meng, and Ermon(2021)}]{song2021denoising}
Song, J.; Meng, C.; and Ermon, S. 2021.
\newblock Denoising Diffusion Implicit Models.
\newblock In \emph{International Conference on Learning Representations (ICLR)}.

\bibitem[{Tang et~al.(2025)Tang, Ayambem, Chuah, and Bharati}]{tang2025perturbation}
Tang, Q.; Ayambem, B.; Chuah, M.~C.; and Bharati, A. 2025.
\newblock Is Perturbation-Based Image Protection Disruptive to Image Editing?
\newblock \emph{2025 IEEE International Conference on Image Processing (ICIP)}.

\bibitem[{Tang and Bharati(2025)}]{tang2025watermarks}
Tang, Q.; and Bharati, A. 2025.
\newblock Watermarks vs. Perturbations for Preventing {AI}-based Style Editing.
\newblock In \emph{The 1st Workshop on GenAI Watermarking}.

\bibitem[{Van~Le et~al.(2023)Van~Le, Phung, Nguyen, Dao, Tran, and Tran}]{van2023anti}
Van~Le, T.; Phung, H.; Nguyen, T.~H.; Dao, Q.; Tran, N.~N.; and Tran, A. 2023.
\newblock Anti-dreambooth: Protecting users from personalized text-to-image synthesis.
\newblock In \emph{IEEE/CVF International Conference on Computer Vision (ICCV)}, 2116--2127.

\bibitem[{Vaswani et~al.(2017)Vaswani, Shazeer, Parmar, Uszkoreit, Jones, Gomez, Kaiser, and Polosukhin}]{vaswani2017attention}
Vaswani, A.; Shazeer, N.; Parmar, N.; Uszkoreit, J.; Jones, L.; Gomez, A.~N.; Kaiser, {\L}.; and Polosukhin, I. 2017.
\newblock Attention is all you need.
\newblock \emph{Advances in neural information processing systems}, 30.

\bibitem[{Wright and Ommer(2022)}]{wright2022artfid}
Wright, M.; and Ommer, B. 2022.
\newblock Artfid: Quantitative evaluation of neural style transfer.
\newblock In \emph{DAGM German Conference on Pattern Recognition}, 560--576. Springer.

\bibitem[{Wynen, Schmid, and Mairal(2018)}]{wynen2018unsupervised}
Wynen, D.; Schmid, C.; and Mairal, J. 2018.
\newblock Unsupervised learning of artistic styles with archetypal style analysis.
\newblock \emph{Advances in Neural Information Processing Systems}, 31.

\bibitem[{Xu et~al.(2024)Xu, Lu, Li, Lu, Wang, and Wei}]{xu2024perturbing}
Xu, J.; Lu, Y.; Li, Y.; Lu, S.; Wang, D.; and Wei, X. 2024.
\newblock Perturbing Attention Gives You More Bang for the Buck: Subtle Imaging Perturbations That Efficiently Fool Customized Diffusion Models.
\newblock In \emph{Proceedings of the IEEE/CVF Conference on Computer Vision and Pattern Recognition (CVPR)}, 24534--24543.

\bibitem[{Xu et~al.(2014)Xu, Tao, Zhang, Wu, and Tsoi}]{xu2014architectural}
Xu, Z.; Tao, D.; Zhang, Y.; Wu, J.; and Tsoi, A.~C. 2014.
\newblock Architectural style classification using multinomial latent logistic regression.
\newblock In \emph{European conference on computer vision (ECCV)}, 600--615. Springer.

\bibitem[{Yao, Li, and Wang(2009)}]{yao2009characterizing}
Yao, L.; Li, J.; and Wang, J.~Z. 2009.
\newblock Characterizing elegance of curves computationally for distinguishing Morrisseau paintings and the imitations.
\newblock In \emph{2009 16th IEEE International Conference on Image Processing (ICIP)}, 73--76. IEEE.

\bibitem[{Zhang et~al.(2018)Zhang, Isola, Efros, Shechtman, and Wang}]{zhang2018unreasonable}
Zhang, R.; Isola, P.; Efros, A.~A.; Shechtman, E.; and Wang, O. 2018.
\newblock The unreasonable effectiveness of deep features as a perceptual metric.
\newblock In \emph{IEEE Conference on Computer Vision and Pattern Recognition (CVPR)}, 586--595.

\bibitem[{Zhang et~al.(2013)Zhang, Cao, Chen, Liu, and Tang}]{zhang2013style}
Zhang, W.; Cao, C.; Chen, S.; Liu, J.; and Tang, X. 2013.
\newblock Style transfer via image component analysis.
\newblock \emph{IEEE Transactions on multimedia}, 15(7): 1594--1601.

\bibitem[{Zhang et~al.(2023)Zhang, Huang, Tang, Huang, Ma, Dong, and Xu}]{zhang2023inversion}
Zhang, Y.; Huang, N.; Tang, F.; Huang, H.; Ma, C.; Dong, W.; and Xu, C. 2023.
\newblock Inversion-based style transfer with diffusion models.
\newblock In \emph{Proceedings of the IEEE/CVF conference on computer vision and pattern recognition (CVPR)}, 10146--10156.

\end{thebibliography}

\end{document}


\maketitle

\section{Details of Dataset}
\label{sec: app_dataset}

\begin{figure}[ht]
    \centering
    \includegraphics[width=\linewidth]{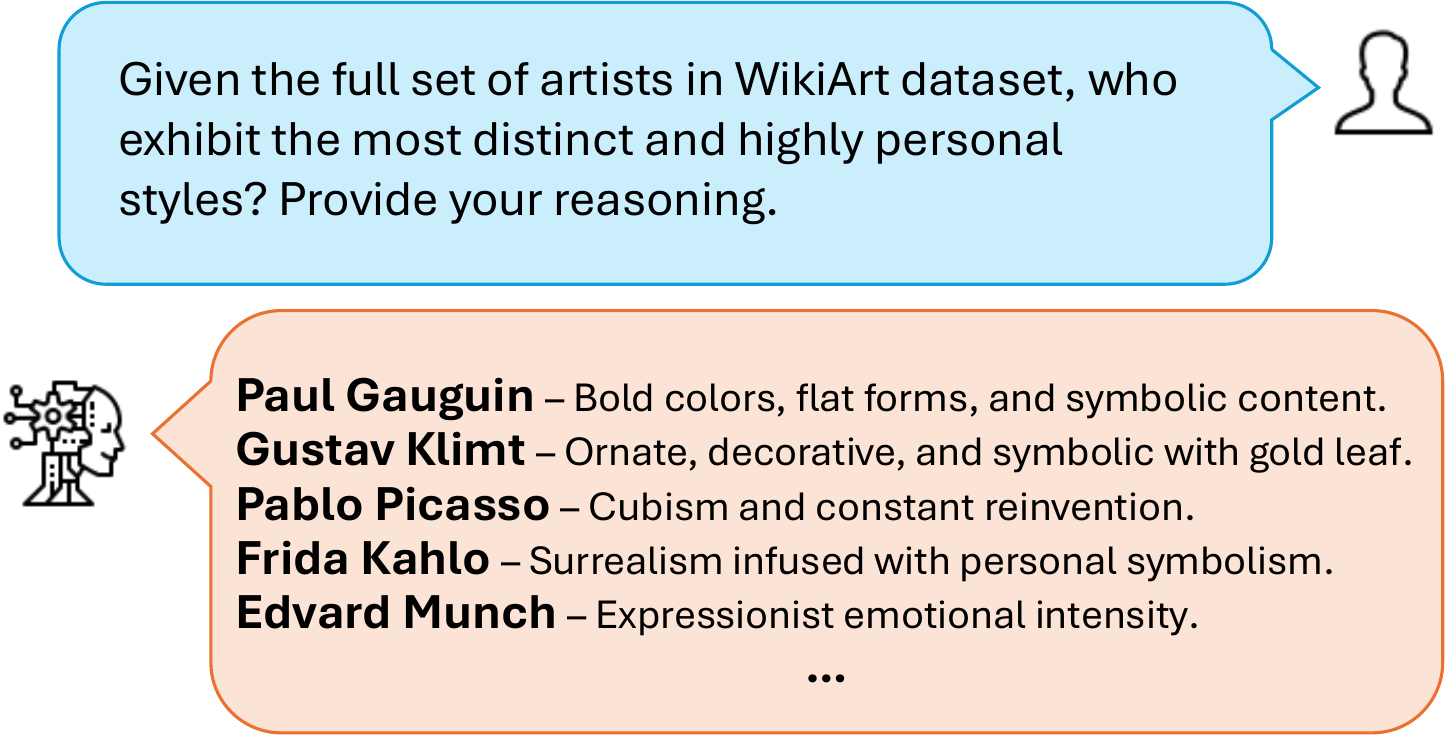}
    \caption{The process of building the refined artwork dataset using ChatGPT~\cite{openai_chatgpt}, for evaluating style protection effectiveness.}
    \label{fig:dataset}
\end{figure}

We provide the detailed description of our refined artwork dataset based on WikiArt~\cite{artgan2018}.
To identify the most stylistically unique artists, we leveraged ChatGPT~\cite{openai_chatgpt} to generate a curated list of renowned artists known for their highly individual and recognizable styles. Figure~\ref{fig:dataset} illustrates the process. Given that each artist possesses a unique style, we represent styles by directly linking them to their respective artist names. Starting with the complete list of artists from the WikiArt dataset~\cite{karras2019style}, we requested the AI assistant to produce a refined selection of artists exhibiting the most distinct and personal styles, along with detailed reasoning, nominating each artist and explaining the distinctive qualities for which they are known. Subsequently, we manually reviewed these nominated candidates and their works to ensure clarity and consistency in stylistic expression. Ultimately, we selected 30 prominent artists whose works demonstrate strong and distinguishable stylistic characteristics. For each artist, representing a distinct style, we curated a small subset of 3 to 5 representative images for customization. This refined dataset forms a compact yet effective benchmark for evaluating style mimicry and protection.

\section{Visualization of Invisibility}

\begin{figure*}[htb]
    \centering
    \includegraphics[width=0.8\linewidth]{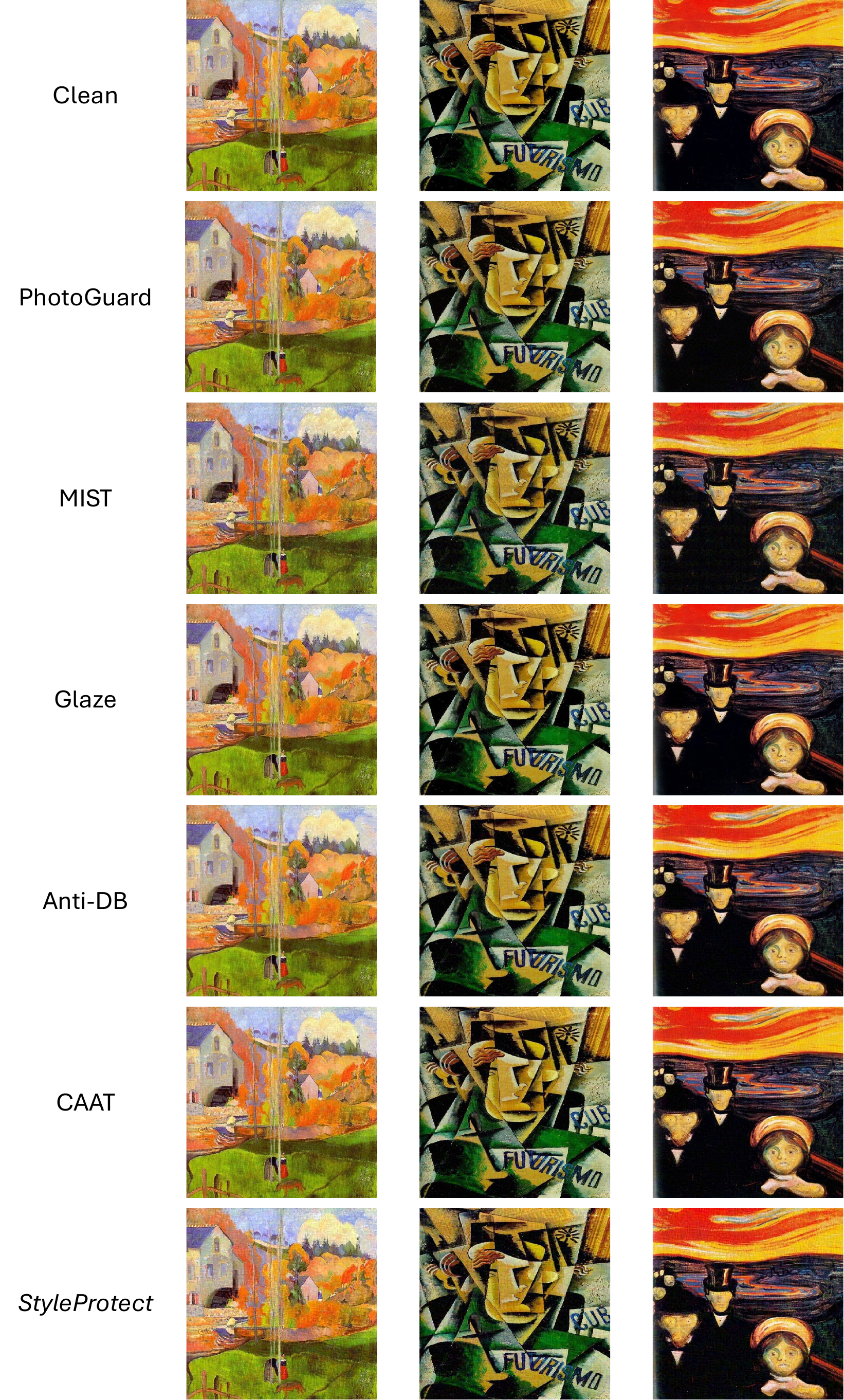}
    \caption{Visualization of perturbations applied by different methods on three artwork images.}
    \label{fig: perturbation}
\end{figure*}

To evaluate the impact of perturbations, Fig.~\ref{fig: perturbation} compares the perturbations applied by different methods.

\section{Details of Style-Sensitive Selection Experiments}

To investigate whether certain cross-attention layers in diffusion models are more sensitive to style representations, we first calculate the average activation strengths to style tokens and content tokens, respectively. Besides, we investigate the alignment between style features extracted from CSD~\cite{somepalli2024measuring} and attention representations from diffusion models. The image is passed through the CSD model, which outputs a style feature embedding of dimension (1×768). In parallel, the image is fed into Stable Diffusion, from which we extract the attention maps corresponding to both style token and content token at each layer. These attention maps have an initial dimensionality of (64×77). To enable a direct comparison between the CSD features and the attention maps, we perform a dimensionality reduction step. Specifically, the (64×77) attention maps are flattened and subsequently projected down to a (1×768) embedding space using linear interpolation. With the features now dimensionally aligned, we compute the cosine similarity between the CSD style embedding and the projected style and content attention maps for each cross-attention layer. This process is repeated for all attention layers. This analysis helps reveal the association between attention activation and style features extracted from CSD.

\bibliography{aaai2026}